\documentclass[conference]{IEEEtran}
\IEEEoverridecommandlockouts
\usepackage{cite}
\usepackage{amsmath,amssymb,amsfonts}
\usepackage{algorithmic,algorithm}
\usepackage{graphicx}
\usepackage{textcomp}
\usepackage{stackengine}
\usepackage{balance}
\usepackage{soul}
\usepackage{xcolor}
\usepackage{booktabs}

\usepackage{caption}

\def\BibTeX{{\rm B\kern-.05em{\sc i\kern-.025em b}\kern-.08em
    T\kern-.1667em\lower.7ex\hbox{E}\kern-.125emX}}

\usepackage{hyperref}
\begin{document}

\title{TIMED: Adversarial and Autoregressive Refinement of Diffusion-Based Time Series Generation}

\author{
    MohammadReza EskandariNasab\IEEEauthorrefmark{1}\thanks{Corresponding author. ORCID: 0009-0004-0697-3716}, 
    Shah Muhammad Hamdi, 
    Soukaina Filali Boubrahimi
    \\
    \textit{School of Computing, Utah State University, Logan, UT 84322, USA} \\
    \{reza.eskandarinasab, s.hamdi, soukaina.boubrahimi\}@usu.edu
}

\maketitle

\begin{abstract}
Generating high-quality synthetic time series is a fundamental yet challenging task across domains such as forecasting and anomaly detection, where real data can be scarce, noisy, or costly to collect. Unlike static data generation, synthesizing time series requires modeling both the marginal distribution of observations and the conditional temporal dependencies that govern sequential dynamics. We propose \textit{TIMED}, a unified generative framework that integrates a denoising diffusion probabilistic model (DDPM) to capture global structure via a forward–reverse diffusion process, a supervisor network trained with teacher forcing to learn autoregressive dependencies through next-step prediction, and a Wasserstein critic that provides adversarial feedback to ensure temporal smoothness and fidelity. To further align the real and synthetic distributions in feature space, \textit{TIMED} incorporates a Maximum Mean Discrepancy (MMD) loss, promoting both diversity and sample quality. All components are built using masked attention architectures optimized for sequence modeling and are trained jointly to effectively capture both unconditional and conditional aspects of time series data. Experimental results across diverse multivariate time series benchmarks demonstrate that \textit{TIMED} generates more realistic and temporally coherent sequences than state-of-the-art generative models.
\end{abstract}


\section{Introduction}
The generation of realistic synthetic time series is a fundamental yet understudied problem in machine learning, especially in domains where data collection is limited by privacy constraints, acquisition costs, or irregular sampling~\cite{giuffre2023harnessing}. These constraints are acute in heliophysics, where multivariate magnetic-field time series are scarce and imbalanced; even with multifaceted preprocessing, including augmentation and contrastive learning~\cite{eskandarinasab2024flare}, performance remains data-limited, motivating synthetic time series augmentation~\cite{Li_2025}. Unlike static modalities such as images \( \mathbf{x} \in \mathbb{R}^{H \times W \times C} \), where \( H \), \( W \), and \( C \) denote height, width, and number of channels, or text sequences \( \mathbf{x} = (w_1, \dots, w_n) \), where \( w_i \) is the \( i \)th token, time series are represented as multivariate sequences \( \mathbf{x}_{1:T} = (x_1, \dots, x_T) \in \mathbb{R}^{T \times F} \), where \( T \) is the sequence length and \( F \) is the number of features. These sequences exhibit complex temporal dependencies, with each \( x_t \) potentially depending on its entire history \( \mathbf{x}_{1:t-1} \)~\cite{bai2018empirical}. These sequences often have variable lengths and follow non-stationary statistics, demanding models capable of learning both intricate temporal dynamics and feature-level distributions. Similar challenges have been investigated in neuroscience, where EEG-based approaches leverage temporal dynamics for tasks such as auditory attention detection~\cite{eskandarinasab2024grucnn} and schizophrenia diagnosis~\cite{raeisi2025eeg}, underscoring the importance of models that capture both short- and long-range dependencies in sequential data. Therefore, a principled generative framework must simultaneously capture the marginal data distribution and the conditional sequential structure. These dual objectives introduce significant modeling challenges, often requiring careful trade-offs between sample realism, temporal coherence, and distributional diversity~\cite{li2022generative}. 

Formally, given a multivariate sequence \( \mathbf{x}_{1:T} \), the generative objective is to approximate the true data-generating distribution \( p(\mathbf{x}_{1:T}) \) with a learned model \( \hat{p}(\mathbf{x}_{1:T}) \). In practical settings, this distribution exhibits intricate structure, characterized by both temporal dependencies and correlations among feature dimensions. Classical sequence models, such as autoregressive and sequence-to-sequence architectures, typically factorize the joint distribution as \( \prod_{t=1}^T p(x_t \mid \mathbf{x}_{1:t-1}) \), enabling effective modeling of causal dynamics. While such models excel in forecasting tasks, they face limitations in generative settings, including exposure bias from teacher forcing~\cite{bengio2015scheduled}, error accumulation during autoregressive sampling, and low sample diversity due to deterministic decoding. These limitations are especially important in applications that require coherent long-horizon dynamics for model-based reasoning, for example equation-editable physics simulations~\cite{jafari2025physmath}.

Recent progress in deep generative models (DGMs) has enabled more holistic approaches to approximating complex data distributions. Prominent DGM families such as Generative Adversarial Networks (GANs)~\cite{goodfellow2014generative}, Variational Autoencoders (VAEs)~\cite{kingma2014auto}, and normalizing flows~\cite{dinh2017density} aim to learn a mapping from latent variables \( \mathbf{z} \sim p(\mathbf{z}) \) to observed data \( \mathbf{x} \sim p(\mathbf{x}) \), but were originally designed for static data and require architectural adaptations to model sequential inputs \( \mathbf{x}_{1:T} \). Denoising Diffusion Probabilistic Models (DDPMs)~\cite{ho2020denoising} define a Markovian forward process \( q(\mathbf{x}^{(\tau)} \mid \mathbf{x}^{(\tau-1)}) \) that progressively adds noise, and learn a reverse process \( p_\theta(\mathbf{x}^{(\tau-1)} \mid \mathbf{x}^{(\tau)}) \) to generate realistic samples~\cite{weng2021diffusion}. While DDPMs have shown impressive performance on image synthesis tasks, their reliance on convolutional architectures optimized for spatial structure limits their effectiveness on time series data, where temporal coherence and autoregressive structure \( p(\mathbf{x}_t \mid \mathbf{x}_{1:t-1}) \) are central. Similar benefits from context-aware sequential modeling appear in NLP, where graph-augmented transformers advance minority stress detection~\cite{chapagain2025transformers}.

To advance the generative modeling of sequential data, \textit{TIMED} proposes a principled framework that addresses key challenges in time series synthesis. Built upon a DDPM backbone, TIMED introduces the following innovations:
\begin{enumerate}
    \item a masked attention DDPM architecture designed to capture global structure in time series;
    \item a supervisor network trained with next-step prediction to enforce autoregressive temporal dependencies;
    \item a Wasserstein critic that provides adversarial guidance for improving temporal realism;
    \item a Maximum Mean Discrepancy (MMD) loss that encourages global distributional alignment between real and generated sequences; and
    \item a unified training strategy that jointly optimizes all three networks, enabling the model to balance fidelity, diversity, and coherence effectively.
\end{enumerate}
Together, these components enable TIMED to generate samples that are both statistically sound and temporally plausible across diverse time series domains.

\section{Related Work}

Generative models differ widely in how they represent data distributions, define training objectives, and minimize divergence between real and generated samples. Autoregressive approaches model the joint distribution via a causal decomposition, \(p(\mathbf{X}_{1:T}) = \prod_{t=1}^T p(X_t \mid \mathbf{X}_{1:t-1})\), and are typically optimized through teacher forcing, which conditions the model on ground-truth observations at each time step~\cite{bengio2015scheduled}. This setup enables exact likelihood computation and stable gradient-based training. However, it introduces a discrepancy between training and inference phases, commonly referred to as exposure bias~\cite{RanzatoCAZ15}: at inference time, the model must autoregressively rely on its own predictions, which can be imperfect, leading to cascading errors and degraded sample quality. Methods such as professor forcing~\cite{lamb2016professor} attempt to mitigate this by aligning the model's hidden-state trajectories across training and generation via adversarial regularization. Despite such remedies, autoregressive models remain deterministic in training and fail to capture latent variability, limiting their capacity as expressive generative models. 

VAEs~\cite{kingma2014auto} learn latent representations through a probabilistic encoder--decoder framework by optimizing the evidence lower bound (ELBO), 
\(\log p(x) \geq \mathbb{E}_{q(z \mid x)}[\log p(x \mid z)] - \mathrm{KL}(q(z \mid x) \| p(z))\)~\cite{kingma2014auto}. 
This formulation promotes a trade-off between reconstruction fidelity and latent regularization. 
While VAEs offer stable training and interpretable structure, they often generate overly smooth or low-variance outputs due to a loose ELBO and the risk of posterior collapse in powerful decoders.

GANs~\cite{goodfellow2014generative} forgo explicit likelihoods by formulating generation as a two-player game between a generator and a discriminator. 
Their training implicitly minimizes the Jensen--Shannon (JS) divergence~\cite{divergence1}:
\[
\min_G \max_D \; \mathbb{E}_{\mathbf{x} \sim p_{\mathrm{data}}}[\log D(\mathbf{x})]
+ \mathbb{E}_{\mathbf{z} \sim p_z}[\log(1 - D(G(\mathbf{z})))].
\]
GANs can produce highly realistic samples, but suffer from training instability, mode collapse, and lack of reliable density estimation due to adversarial dynamics and sensitivity to hyperparameters. TimeGAN~\cite{timegan} is a hybrid framework for time series generation that combines GANs with supervised autoregressive learning. It jointly trains a generator, discriminator, embedding network, and a supervised loss module using a combination of adversarial, supervised, and reconstruction losses. By generating sequences in a learned embedding space rather than directly in the feature space, TimeGAN mitigates issues such as mode collapse and training instability, resulting in more coherent and temporally consistent synthetic data. ChronoGAN~\cite{eskandarinasab2025chronogan} extends TimeGAN by training the generator in a learned latent space while the discriminator operates in feature space, and introduces an early-generation schedule. SeriesGAN~\cite{eskandarinasab2024seriesgan} further advances ChronoGAN by employing dual discriminators operating in latent and feature spaces, improving training stability and sample quality.

Adversarial Autoencoders (AAEs)~\cite{adversarialautoencoders} combine reconstruction-based objectives with adversarial training to impose structure on the latent space. 
Instead of using the Kullback--Leibler (KL) divergence~\cite{divergence2} as in VAEs, AAEs match the aggregated posterior \(q(z) = \int q(z \mid x) p(x)\,dx\) to a predefined prior \(p(z)\) via an adversarial loss that minimizes \(\mathrm{JS}(q(z) \| p(z))\). 
This strategy enables flexible prior design and promotes more structured latent representations. 
However, AAEs inherit the instability of adversarial training and lack a tractable likelihood, complicating quantitative evaluation. AVATAR~\cite{avatar} is a time series generation framework based on AAEs, designed to capture both global data structure and local temporal dynamics. It integrates a supervisory autoregressive network to model sequence dependencies, a novel distribution loss to enhance latent space regularization, and a GRU-based decoder with batch normalization for improved stability. Through joint training of all components, AVATAR achieves high-quality, diverse, and temporally coherent synthetic sequences.

Flow-based models~\cite{dinh2014nice, dinh2017density, kingma2018glow} provide a likelihood-based framework by employing a sequence of invertible transformations with tractable Jacobians. The log-density of an input \(x\) is evaluated using the change-of-variables formula:
\[
\log p_\theta(x) = \log p_Z(f_\theta(x)) + \log \left| \det \left( \frac{\partial f_\theta(x)}{\partial x} \right) \right|.
\]
These models enable exact likelihood computation and efficient sampling. However, their expressiveness is constrained by the requirement for invertibility and the necessity of maintaining computationally tractable Jacobians, which can limit performance on complex, multimodal data distributions.

DDPMs~\cite{ho2020denoising} extend likelihood-based generative modeling by defining the data generation process as the reverse of a fixed diffusion trajectory, in which Gaussian noise is progressively added over a sequence of latent variables \( \{\mathbf{x}^{(\tau)}\}_{\tau=1}^{T_{\text{diff}}} \). Inspired by thermodynamic principles, this formulation constructs a Markovian noising process \(q(\{\mathbf{x}^{(\tau)}\}_{\tau=1}^{T_{\text{diff}}} \mid \mathbf{x}^{(0)})\), where each step adds Gaussian noise with a predefined variance schedule:
\[
q(\mathbf{x}^{(\tau)} \mid \mathbf{x}^{(\tau-1)}) = \mathcal{N}(\mathbf{x}^{(\tau)}; \sqrt{1 - \beta_\tau} \mathbf{x}^{(\tau-1)}, \beta_\tau \mathbf{I}).
\]
The generative process learns to reverse this trajectory by modeling the conditional transitions \(p_\theta(\mathbf{x}^{(\tau-1)} \mid \mathbf{x}^{(\tau)})\). Training reduces to minimizing a weighted mean squared error between the predicted and true noise, providing a stable and interpretable objective. While DDPMs achieve strong mode coverage and high-quality sample generation without adversarial training, their reliance on sequential denoising steps leads to significant inference latency~\cite{ddim}. Furthermore, the use of convolutional backbones hinders their ability to capture complex temporal dependencies in time series data.

DiffTime~\cite{difftime} is a diffusion-based generative model tailored for multivariate time series synthesis. It builds on the DDPM framework by introducing a temporal-aware U-Net architecture that captures local and global temporal dependencies at multiple scales. DiffTime incorporates sinusoidal time embeddings to condition the model on the diffusion timestep, and supports conditional generation via auxiliary inputs such as class labels. The training objective minimizes the denoising score matching loss, guiding the model to recover clean sequences from noisy inputs.

Our proposed framework \textit{TIMED} addresses the limitations of DDPMs on time series by replacing convolutional backbones with masked attention layers better suited for sequential data. It further augments the diffusion model with a supervisor for autoregressive structure and a Wasserstein critic for temporal smoothness, enabling both statistical fidelity and temporal coherence.

\section{Problem Formulation}
Let \(\mathcal{X}\) represent the domain of temporal attributes, and let \(X \in \mathcal{X}\) denote a random vector whose realizations are written as \(x\). Our analysis centers on sequences of such vectors,
\[
\mathbf{X}_{1:T_{\text{data}}} = (X_1, X_2, \dots, X_{T_{\text{data}}}),
\]
where \(T_{\text{data}}\) denotes the sequence length. For notational convenience, we refer to it simply as \(T\) throughout this section.

These sequences are drawn from an underlying joint distribution \(p(\mathbf{X}_{1:T})\).  The training dataset comprises \(N\) independent sequences of varying lengths,
\[
\mathcal{D} = \bigl\{\mathbf{X}_{1:T^{(n)}}^{(n)}\bigr\}_{n=1}^N.
\]
Likewise, we omit the instance index \(n\) when it is clear from context.
The primary goal of DDPMs is to use \(\mathcal{D}\) to construct an estimated distribution \(\hat{p}(\mathbf{X}_{1:T})\) that closely mirrors the true data distribution \(p(\mathbf{X}_{1:T})\). This is challenging due to the high dimensionality of each \(\mathbf{X}_{1:T}\), the potential for complex behavior, and the need to capture both short- and long-range temporal dependencies. To make learning tractable, we adopt an autoregressive factorization of the joint distribution:
\[
p(\mathbf{X}_{1:T}) = \prod_{t=1}^{T} p\bigl(X_t \mid \mathbf{X}_{1:t-1}\bigr).
\]
This reformulation focuses the problem on learning a sequence of conditional distributions \(\hat{p}(X_t \mid \mathbf{X}_{1:t-1})\) that approximate the true conditionals \(p(X_t \mid \mathbf{X}_{1:t-1})\) at each time step. Consequently, our learning process is driven by two complementary objectives~\cite{timegan}:

\begin{enumerate}
  \item \textbf{Sequence‐level (global) objective:}  
    Aligns the joint distribution of sequences between real and generated data:
    \[
    \min_{\hat{p}} D\left( p(\mathbf{X}_{1:T}) \,\|\, \hat{p}(\mathbf{X}_{1:T}) \right),
    \]
    where \(D\) denotes a suitable divergence measure.

  \item \textbf{Step‐wise (local) objective:}  
    Matches conditional distributions at each timestep \(t\):
    \[
    \min_{\hat{p}} D\left( p(X_t \mid \mathbf{X}_{1:t-1}) \,\|\, \hat{p}(X_t \mid \mathbf{X}_{1:t-1}) \right).
    \]
\end{enumerate}

By combining these objectives, our framework captures both the global structure of entire sequences and the local dynamics at each timestep, yielding models that are robust to mode collapse and capable of precise temporal forecasting.

\section{Proposed Method: TIMED}

\begin{figure*}
\centering
\includegraphics[width=0.7\textwidth]{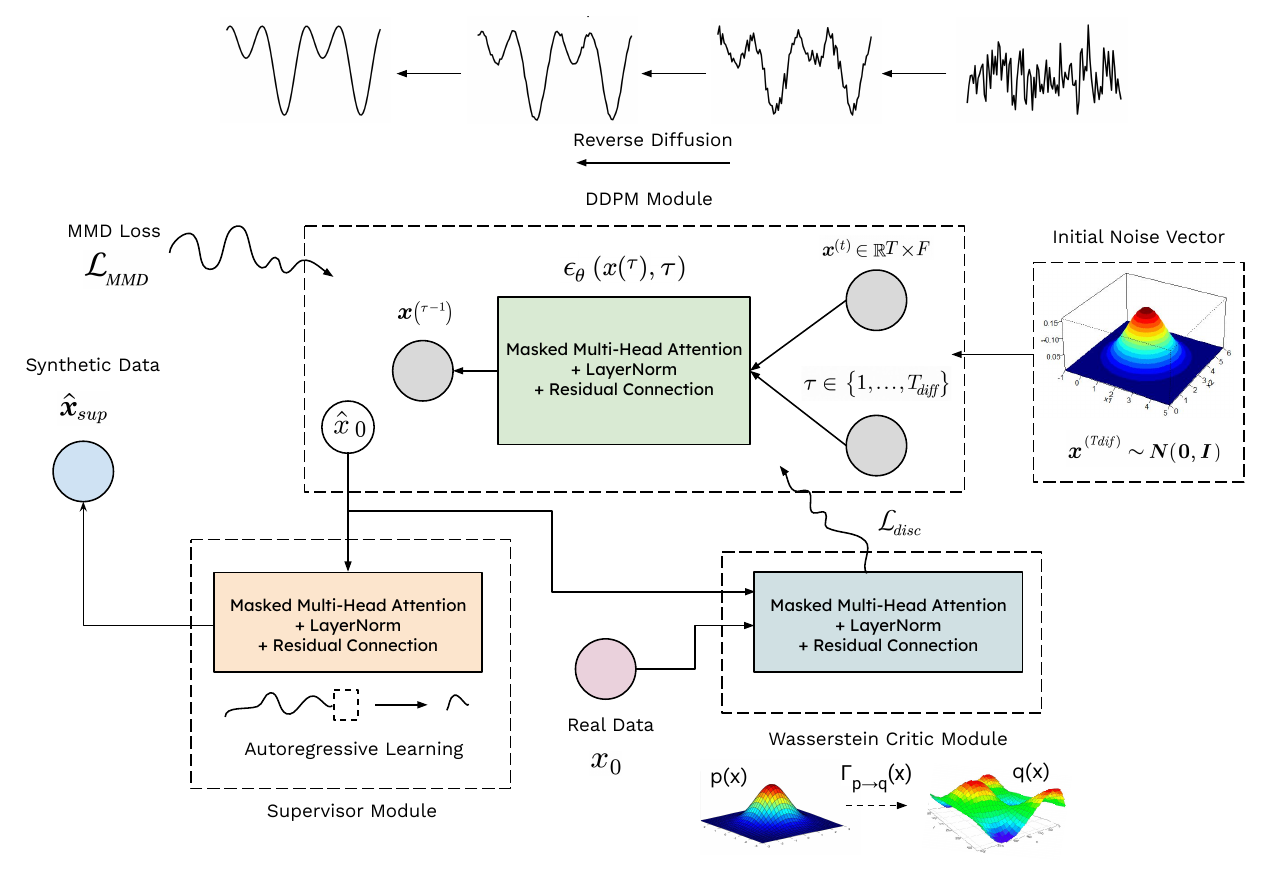}
\caption{Schematic of the \textit{TIMED} framework. Starting from Gaussian noise \( \mathbf{x}^{(T_{\text{diff}})} \sim \mathcal{N}(0, \mathbf{I}) \), the DDPM module denoises the sample using masked Transformer layers to generate \( \hat{\mathbf{x}}_0 \). This is refined by the Supervisor Module to produce the final output \( \hat{\mathbf{x}}_{\text{sup}} \). A Wasserstein Critic and MMD loss enforce distributional alignment with real data \( \mathbf{x}_0 \). All components share a masked attention architecture with residual connections and LayerNorm.}
\label{fig:timed}
\end{figure*}

The architectural details and design of the \textit{TIMED} framework are illustrated in Fig.~\ref{fig:timed}, which depicts the overall information flow between modules, highlighting the roles of diffusion steps, autoregressive supervision, and distributional alignment via the Wasserstein critic and MMD loss.

\subsection{Diffusion Models with Masked Transformers}

DDPMs~\cite{ho2020denoising, sohl2015deep, yang2023diffusion} constitute a powerful family of generative models that gradually corrupt observed data using Gaussian noise in a forward diffusion process and learn to reverse this degradation through a neural network. In the context of \textit{TIMED}, let \( \mathbf{x}_0 \in \mathbb{R}^{T_{\text{data}} \times F} \) denote a multivariate time series with \( T_{\text{data}} \) time steps and \( F \) features. The forward process initializes with \( \mathbf{x}^{(0)} = \mathbf{x}_0 \) and defines a Markov chain over latent variables \( \{ \mathbf{x}^{(\tau)} \}_{\tau=1}^{T_{\text{diff}}} \) via the recursive addition of Gaussian noise:
\begin{equation}
\begin{aligned}
    q(\mathbf{x}^{(\tau)} \mid \mathbf{x}^{(\tau-1)}) 
    &= \mathcal{N}\left( \mathbf{x}^{(\tau)}; \sqrt{1 - \beta_\tau} \mathbf{x}^{(\tau-1)}, \beta_\tau \mathbf{I} \right), \\
    &\quad \tau = 1, \dots, T_{\text{diff}}.
\end{aligned}
\end{equation}
where each \( \beta_\tau \in (0, 1) \) controls the noise magnitude at timestep \( \tau \), and \( \mathbf{I} \) denotes the identity matrix, implying independent noise across all feature dimensions. We employ a linear noise schedule for simplicity in our implementation. The notation \( \mathcal{N}(\cdot; \mu, \Sigma) \) denotes a multivariate Gaussian with mean \( \mu \) and covariance \( \Sigma \).

This formulation admits a closed-form marginal distribution that allows directly sampling noisy inputs at arbitrary diffusion steps without explicitly simulating all previous transitions:
\begin{equation}
    q(\mathbf{x}^{(\tau)} \mid \mathbf{x}_0) = \mathcal{N}\left( \mathbf{x}^{(\tau)}; \sqrt{\bar{\alpha}_\tau} \mathbf{x}_0, (1 - \bar{\alpha}_\tau) \mathbf{I} \right),
\end{equation}
where \( \bar{\alpha}_\tau = \prod_{s=1}^\tau (1 - \beta_s) \) denotes the cumulative noise decay factor. During training, this closed-form is used to efficiently construct noisy inputs from clean data at randomly sampled steps \( \tau \) using:
\begin{equation}
\mathbf{x}^{(\tau)} = \sqrt{\bar{\alpha}_\tau} \mathbf{x}_0 + \sqrt{1 - \bar{\alpha}_\tau} \boldsymbol{\epsilon}, \quad \boldsymbol{\epsilon} \sim \mathcal{N}(0, \mathbf{I}).
\end{equation}

To reverse the diffusion process, a neural network is trained to approximate the conditional distribution of the previous step:
\begin{equation}
    p_\theta(\mathbf{x}^{(\tau-1)} \mid \mathbf{x}^{(\tau)}) = \mathcal{N}\left( \mathbf{x}^{(\tau-1)}; \boldsymbol{\mu}_\theta(\mathbf{x}^{(\tau)}, \tau), \sigma_\tau^2 \mathbf{I} \right),
\end{equation}
where \( \boldsymbol{\mu}_\theta \) is a learnable mean function, and \( \sigma_\tau^2 \) is the variance, typically derived analytically. Following~\cite{ho2020denoising}, the mean is expressed in terms of a noise prediction network \( \boldsymbol{\epsilon}_\theta \) as:
\begin{equation}
    \boldsymbol{\mu}_\theta(\mathbf{x}^{(\tau)}, \tau) = \frac{1}{\sqrt{1 - \beta_\tau}} \left( \mathbf{x}^{(\tau)} - \frac{\beta_\tau}{\sqrt{1 - \bar{\alpha}_\tau}} \boldsymbol{\epsilon}_\theta(\mathbf{x}^{(\tau)}, \tau) \right).
\end{equation}

The reverse-step variance is given by:
\begin{equation}
    \sigma_\tau^2 = \tilde{\beta}_\tau = \frac{1 - \bar{\alpha}_{\tau-1}}{1 - \bar{\alpha}_\tau} \beta_\tau,
\end{equation}
which captures posterior uncertainty. Instead of explicitly modeling the full posterior \( q(\mathbf{x}^{(\tau-1)} \mid \mathbf{x}^{(\tau)}, \mathbf{x}_0) \), training minimizes a denoising loss between true and predicted noise:
\begin{equation}
    \mathcal{L}_{\text{DDPM}} = \mathbb{E}_{\mathbf{x}_0, \tau, \boldsymbol{\epsilon}} \left[ \left\| \boldsymbol{\epsilon} - \boldsymbol{\epsilon}_\theta \left( \sqrt{\bar{\alpha}_\tau} \mathbf{x}_0 + \sqrt{1 - \bar{\alpha}_\tau} \boldsymbol{\epsilon}, \tau \right) \right\|_2^2 \right].
\end{equation}

During inference, sampling begins from isotropic Gaussian noise \( \mathbf{x}^{(T_{\text{diff}})} \sim \mathcal{N}(0, \mathbf{I}) \), and the learned reverse process is iteratively applied to generate synthetic samples \( \hat{\mathbf{x}}_0 \).

\vspace{0.6em}
\noindent
\textbf{Architectural Design in \textit{TIMED}.} In contrast to conventional DDPMs that rely on convolutional backbones, \textit{TIMED} leverages a masked Transformer architecture optimized for temporal data. Given a noisy input \( \mathbf{x}^{(\tau)} \in \mathbb{R}^{T_{\text{data}} \times F} \), the model builds its internal representation as:
\begin{align}
    \mathbf{h}^{(0)} ={} & \ \textit{EmbedProj}(\mathbf{x}^{(\tau)}) 
    + \textit{TimeEmbedding}(\tau) \notag \\
    & + \textit{PositionalEmbedding}(T_{\text{data}}),
\end{align}
where \textit{EmbedProj} denotes a linear projection of the input features, \textit{TimeEmbedding} represents a learnable encoding of the current diffusion step, and \textit{PositionalEmbedding} captures the temporal order of the sequence.

The latent representation \( \mathbf{h}^{(0)} \in \mathbb{R}^{T_{\text{data}} \times d} \) is passed through a stack of \( L \) masked Transformer layers:
\begin{equation}
    \mathbf{h}^{(\ell)} = \text{MaskedTransformerBlock}(\mathbf{h}^{(\ell-1)}), \quad \ell = 1, \dots, L,
\end{equation}
Each block contains causal self-attention and feed-forward layers. Specifically, the attention mechanism computes~\cite{vaswani2017attention}:
\begin{equation}
    \text{Attention}(Q, K, V) = \text{softmax}\left( \frac{QK^\top}{\sqrt{d_k}} + \mathbf{M} \right)V,
\end{equation}
where \( Q = \mathbf{h}W^Q \), \( K = \mathbf{h}W^K \), and \( V = \mathbf{h}W^V \) are the query, key, and value matrices with learned projections \( W^Q, W^K, W^V \in \mathbb{R}^{d \times d_k} \). The mask matrix \( \mathbf{M} \in \mathbb{R}^{T_{\text{data}} \times T_{\text{data}}} \) assigns \( -\infty \) to positions that violate causal ordering, ensuring each timestep attends only to its past. Following attention, the output undergoes position-wise feed-forward transformation:
\begin{equation}
    \text{FFN}(\mathbf{h}) = \left( \text{ReLU}(\mathbf{h}W_1 + b_1) \right) W_2 + b_2,
\end{equation}
with learned parameters \( W_1 \in \mathbb{R}^{d \times d_{\text{ff}}} \), \( W_2 \in \mathbb{R}^{d_{\text{ff}} \times d} \), and biases \( b_1, b_2 \). Residual connections and layer normalization are applied after each sub-layer to improve training stability.

The final noise estimate is produced by mapping the output of the last Transformer block through a linear output head:
\begin{equation}
    \boldsymbol{\epsilon}_\theta(\mathbf{x}^{(\tau)}, \tau) = \text{NoiseHead}(\mathbf{h}^{(L)}),
\end{equation}
where \( \text{NoiseHead} \) is a fully connected layer that maps each \( d \)-dimensional latent vector to the original feature space.

This architecture enables \textit{TIMED} to model long-range temporal dependencies, enforce causal generation, and produce diverse, high-fidelity time series samples.

\subsection{Autoregressive Supervised Learning}
While diffusion models excel at learning complex data distributions through iterative denoising, they lack explicit enforcement of temporal dynamics inherent to time series. To address this, TIMED incorporates an \text{autoregressive supervised learning} (ASL) module that directly models the temporal progression of the data, encouraging consistency and causal dependency across time steps.

Let $\hat{\mathbf{x}}_0 = \boldsymbol{\epsilon}_\theta(\mathbf{x}^{(\tau)}, \tau)$ denote the denoised estimate produced by the main DDPM module. We introduce a causal decoder $f_{\text{AR}}$, which shares the same masked Transformer architecture as the main denoising model. This shared architecture includes time and positional embeddings, causal attention layers, and layer-normalized residual blocks designed to preserve the autoregressive structure of sequences.

The objective of $f_{\text{AR}}$ is to learn a conditional distribution over future values given the past denoised inputs. Specifically, we minimize the autoregressive loss:
\begin{equation}
\mathcal{L}_{\text{AR}} = \frac{1}{T - \Delta} \sum_{t=1}^{T - \Delta} \left\| \mathbf{x}_t - f_{\text{AR}}(\hat{\mathbf{x}}_{1:t-\Delta}) \right\|^2_2,
\end{equation}
where $\Delta$ denotes a small lookahead gap, and $\hat{\mathbf{x}}_{1:t-\Delta}$ is the predicted history from which the future point $\mathbf{x}_t$ is forecasted. The loss enforces temporal consistency by encouraging the model to reconstruct or predict values based on prior information only, thus maintaining a strict autoregressive flow.

The masked attention mechanism within $f_{\text{AR}}$ ensures that, at each time step $t$, the model has access only to $\hat{\mathbf{x}}_1, \dots, \hat{\mathbf{x}}_{t-1}$. This strictly causal design prevents information leakage from future steps, allowing the network to learn the natural sequential dependencies present in real-world temporal data. The model captures both short- and long-range dependencies, enabling robust learning of complex dynamics such as periodicity, seasonality, and trend shifts.

Finally, the synthetic time series data output by TIMED is not taken directly from the DDPM decoder. Instead, the final generation step is delegated to the autoregressive supervisor $f_{\text{AR}}$, which refines the denoised output $\hat{\mathbf{x}}_0$ into temporally coherent and high-fidelity sequences:
\begin{equation}
\hat{\mathbf{x}}_{\text{synth}} = f_{\text{AR}}(\hat{\mathbf{x}}_0).
\end{equation}

This design ensures that the generative model not only captures the distributional characteristics of the data (via DDPM) but also adheres to its temporal structure (via ASL), leading to improved realism and predictive utility in synthetic sequences.

\vspace{0.5em}
\subsection{Wasserstein Critic for Distribution Matching}
Although DDPMs are highly effective at modeling local noise distributions, they do not directly enforce global distribution alignment between real and generated sequences. To bridge this gap, TIMED introduces a \textit{Wasserstein critic} $D_\phi$ to explicitly measure and minimize the divergence between the empirical data distribution $p_{\text{data}}$ and the model distribution $p_g$ of synthetic samples.

The Wasserstein-1 distance~\cite{wgan} (also called Earth Mover's Distance) between two probability distributions \( p \) and \( q \) over metric space \( \mathcal{X} \) is defined as:
\begin{equation}
W(p, q) = \inf_{\gamma \in \Pi(p, q)} \mathbb{E}_{(x, y) \sim \gamma}[\|x - y\|],
\end{equation}
where \( \Pi(p, q) \) denotes the set of all joint distributions with marginals \( p \) and \( q \). This quantity reflects the minimum cost of transforming one distribution into the other, and unlike other divergences (e.g., KL or JS), it remains well-defined and smooth even when the distributions do not overlap.

To approximate this distance efficiently, we employ the Kantorovich–Rubinstein duality~\cite{villani2008optimal}, which allows the Wasserstein distance to be expressed as:
\begin{equation}
W(p, q) = \sup_{\|f\|_L \leq 1} \mathbb{E}_{x \sim p}[f(x)] - \mathbb{E}_{y \sim q}[f(y)],
\end{equation}
where the supremum is taken over all 1-Lipschitz functions \( f \). In our setup, \( f \) is parameterized by a neural network \( D_\phi \), referred to as the \textit{critic}. We train the critic using the improved WGAN-GP~\cite{gulrajani2017improved} objective:
\begin{align}
\mathcal{L}_{\text{W}} &= \mathbb{E}_{\hat{\mathbf{x}} \sim p_g} [ D_\phi(\hat{\mathbf{x}}) ] 
- \mathbb{E}_{\mathbf{x} \sim p_{\text{data}}} [ D_\phi(\mathbf{x}) ] \nonumber \\
&\quad + \lambda \, \mathbb{E}_{\tilde{\mathbf{x}} \sim p_{\tilde{\mathbf{x}}}} 
\left[ \left( \left\| \nabla_{\tilde{\mathbf{x}}} D_\phi(\tilde{\mathbf{x}}) \right\|_2 - 1 \right)^2 \right],
\end{align}
where \( \tilde{\mathbf{x}} = \alpha \mathbf{x} + (1 - \alpha)\hat{\mathbf{x}} \), for \( \alpha \sim \text{Uniform}(0, 1) \), is a randomly interpolated sample between real and generated data, and \( \lambda \) is a gradient penalty coefficient (typically set to 10). The gradient penalty enforces the Lipschitz constraint by penalizing deviations of the critic's gradient norm from 1.

The Wasserstein loss is integrated with the denoising and supervised losses to jointly guide the training.
This multi-objective design helps TIMED achieve sharper alignment with the real data distribution while preserving temporal consistency and diversity.

\subsection{Maximum Mean Discrepancy Loss}

Although DDPMs are effective at producing high-quality samples by minimizing denoising-based reconstruction losses, they do not inherently ensure that the generated and real data distributions are statistically aligned. To address this, we introduce the MMD loss, which quantifies the divergence between distributions using a kernel-based metric~\cite{mmd2012}. 

Let $\{\mathbf{x}_i\}_{i=1}^n \sim p$ and $\{\mathbf{y}_j\}_{j=1}^n \sim q$ denote two sets of samples drawn from distributions \(p\) and \(q\), respectively. The empirical squared MMD using a radial basis function (RBF) kernel \(k(\mathbf{a}, \mathbf{b}) = \exp\left(-\|\mathbf{a} - \mathbf{b}\|^2 / 2\sigma^2\right)\) is given by:

\begin{align}
\mathrm{MMD}^2(p, q)
&= \frac{1}{n(n-1)} \sum_{i \ne j} k(\mathbf{x}_i, \mathbf{x}_j) \nonumber \\
&\quad + \frac{1}{n(n-1)} \sum_{i \ne j} k(\mathbf{y}_i, \mathbf{y}_j)
- \frac{2}{n^2} \sum_{i, j} k(\mathbf{x}_i, \mathbf{y}_j).
\end{align}

To compute this loss on multivariate time series, we reshape each sample \(\mathbf{x} \in \mathbb{R}^{T \times F}\) into a flat vector \(\mathbf{x} \in \mathbb{R}^{d}\), where \(d = T \cdot F\). The resulting MMD loss is:

\begin{equation}
\mathcal{L}_{\text{MMD}} = \mathrm{MMD}^2(\{\mathbf{x}_i\}_{i=1}^n, \{\hat{\mathbf{x}}_i\}_{i=1}^n).
\end{equation}

By minimizing this term, the model is encouraged to match higher-order statistical properties between the generated and real data, thereby improving the global distributional accuracy of the synthetic sequences.

\subsection{Joint Training}

The training process of the \textit{TIMED} framework is carried out in three stages to ensure stability and effective convergence of its constituent modules.

\paragraph{Stage 1. Supervised Pretraining}  
We begin by pretraining the autoregressive supervisor \( f_{\text{AR}} \), which learns to predict future segments of a sequence given a denoised prefix. Given the ground truth input \( \mathbf{x}_0 \in \mathbb{R}^{T \times F} \), a masking policy hides the last \( \Delta \) time steps. The supervisor is trained to minimize the forecasting loss:
\begin{equation}
    \mathcal{L}_{\text{AR}} = \frac{1}{\Delta} \sum_{t = T - \Delta + 1}^{T} \left\| \hat{\mathbf{x}}_t - \mathbf{x}_t \right\|_2^2,
\end{equation}
where \( \hat{\mathbf{x}}_t = f_{\text{AR}}(\mathbf{x}_{1:t-1}) \) and \( f_{\text{AR}} \) shares the same masked Transformer architecture used in the diffusion module to preserve causal attention and temporal dependencies.

\paragraph{Stage 2. Diffusion Pretraining}  
Next, we train the DDPM module using the standard denoising loss:
\begin{equation}
    \mathcal{L}_{\text{DDPM}} = \mathbb{E}_{\mathbf{x}_0, \tau, \boldsymbol{\epsilon}} \left[ \left\| \boldsymbol{\epsilon} - \boldsymbol{\epsilon}_\theta \left( \sqrt{\bar{\alpha}_\tau} \mathbf{x}_0 + \sqrt{1 - \bar{\alpha}_\tau} \boldsymbol{\epsilon}, \tau \right) \right\|_2^2 \right],
\end{equation}
where \( \boldsymbol{\epsilon}_\theta \) denotes the noise prediction network parameterized by the masked Transformer.

\paragraph{Stage 3: Joint Optimization.}  
In the final stage, all modules, diffusion, supervisor, and Wasserstein critic, are trained together under a unified objective. We introduce the total loss function:
\begin{equation}
    \mathcal{L}_{\text{TIMED}} = \mathcal{L}_{\text{DDPM}} + \lambda_{\text{AR}} \mathcal{L}_{\text{AR}} + \lambda_{\text{MMD}} \mathcal{L}_{\text{MMD}} + \lambda_{\text{W}} \mathcal{L}_{\text{W}},
\end{equation}
where:
\begin{itemize}
    \item \( \mathcal{L}_{\text{DDPM}} \) is the denoising loss from the diffusion module,
    \item \( \mathcal{L}_{\text{AR}} \) is the supervised forecasting loss,
    \item \( \mathcal{L}_{\text{MMD}} \) enforces distributional alignment between real and synthetic samples using Maximum Mean Discrepancy,
    \item \( \mathcal{L}_{\text{W}} \) is the Wasserstein critic loss with gradient penalty,
    \item and \( \lambda_{\text{AR}}, \lambda_{\text{MMD}}, \lambda_{\text{W}} \) are tunable hyperparameters to balance each term.
\end{itemize}

This joint training strategy ensures that the generator produces temporally consistent samples (via \( \mathcal{L}_{\text{AR}} \)), globally aligned with real data distributions (via \( \mathcal{L}_{\text{MMD}} \)), while being guided by the stable gradients of the Wasserstein critic (via \( \mathcal{L}_{\text{W}} \)). The final synthetic samples are decoded from the autoregressive supervisor applied to the denoised outputs of the diffusion process.

\section{Experiments and Discussion}
An open-source implementation of \textit{TIMED} using TensorFlow and Keras, along with all datasets and configuration files, is available in an anonymous repository\footnote{Code and data: \url{https://github.com/samresume/TIMED}} for reproducibility.

\textit{TIMED} is evaluated against a diverse set of state-of-the-art baselines, including \textbf{DiffTime}~\cite{difftime}, \textbf{Vanilla DDPM}~\cite{ho2020denoising}, \textbf{AVATAR}~\cite{avatar}, \textbf{TimeGAN}~\cite{timegan}, and \textbf{Teacher Forcing (T-Forcing)}~\cite{bengio2015scheduled}. These methods represent a range of generative modeling paradigms, encompassing diffusion-based, adversarial autoencoder-based, GAN-based, and autoregressive approaches. To ensure a fair and meaningful comparison, we adopt the optimal hyperparameter configurations as reported in the original works for each respective baseline.

\subsection{Datasets}
\begin{figure}
\centering
\includegraphics[width=0.5\textwidth]{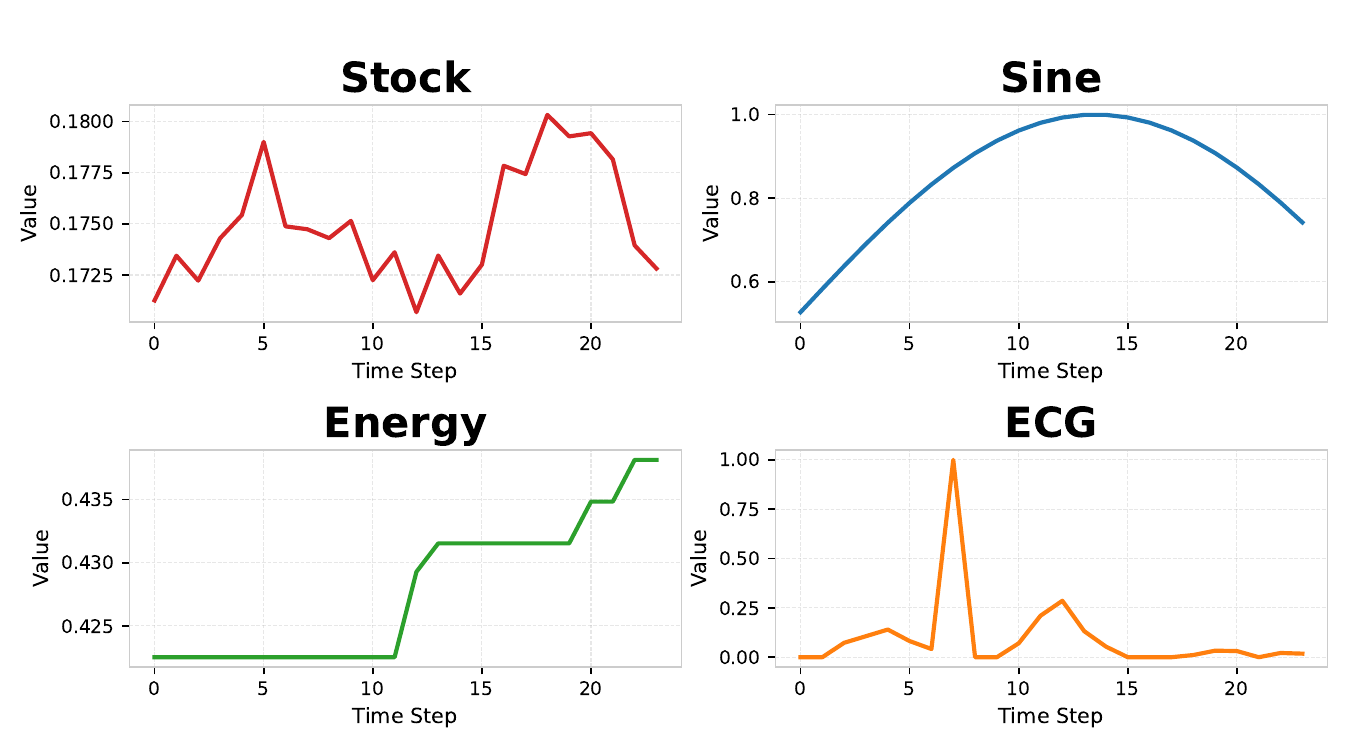}
\caption{The figure presents four randomly selected samples, one from each of the datasets used in this study.}
\label{fig:datasets}
\end{figure}

To evaluate the effectiveness of the proposed \textit{TIMED} framework, we conduct experiments on a diverse set of benchmark time series datasets, as summarized in Fig.~\ref{fig:datasets} and Table~\ref{tab:dataset_stats}. These datasets span both synthetic and real-world domains, enabling a comprehensive assessment of \textit{TIMED}'s robustness and generalizability.

\begin{table}[h]
\centering
\caption{Summary statistics for the time series datasets used in evaluation.}
\begin{tabular}{lccc}
\toprule
\textbf{Dataset} & \textbf{\# Instances} & \textbf{Sequence Length} & \textbf{\# Features} \\
\midrule
Stocks  & 4,407  & 24 & 6 \\
Sines   & 10,000 & 24 & 4 \\
Energy  & 19,711 & 24 & 5 \\
ECG   & 15,000 & 24 & 3 \\
\bottomrule
\end{tabular}
\label{tab:dataset_stats}
\end{table}

\begin{enumerate}
    \item \textbf{Sines.} Synthetic multivariate sequences generated using \( x_i(t) = 0.5 \cdot (\sin(\eta t + \theta) + 1) \), with \(\eta \sim U[0.1, 0.2]\) and \(\theta \sim U[0, 0.1] \). The transformation bounds values in \([0, 1]\), simulating simple periodic dynamics.

    \item \textbf{ECG.} Synthetic electrocardiogram-like sequences composed of one or two heartbeat patterns per sample, with randomized positions and morphological variability in P, Q, R, S, and T waves. Each component’s amplitude and width are sampled from wide distributions, and Gaussian noise is added per feature.

    \item \textbf{Stocks.} Real-world Google stock data (2004--2022) including six indicators: trading volume, daily high/low, opening, closing, and adjusted closing prices. Sequences are continuous, noisy, and shaped by market dynamics.

    \item \textbf{Energy.} A filtered subset of the UCI Appliances Energy Prediction dataset~\cite{UCI_Appliances_Energy}, containing multivariate time series with noisy periodicity and inter-feature correlations. Five features related to energy consumption and environmental factors are used.
\end{enumerate}

\subsection{Evaluation Metrics}
Evaluating generative models for time series requires a comprehensive framework that considers fidelity, diversity, and practical utility. As a result, the literature commonly adopts three core metrics: discriminative score, predictive score, and distributional alignment, which are now standard in evaluating time series generation models.

\begin{table*}
\centering
\caption{Discriminative and predictive scores for time series generation methods on four benchmark datasets. Lower values indicate better performance.}
\label{tbl:scores}
\small
\begin{tabular}{lcccc}
\toprule
\multicolumn{5}{c}{\textit{Discriminative Score}} \\
\midrule
\textbf{Method} & \textbf{Stocks} & \textbf{Sines} & \textbf{ECG} & \textbf{Energy} \\
\midrule
\textbf{TIMED} & \textbf{0.093 $\pm$ 0.00} & \underline{0.171 $\pm$ 0.02} & \textbf{0.034 $\pm$ 0.00} & \textbf{0.105 $\pm$ 0.01} \\
DiffTime        & \underline{0.129 $\pm$ 0.00} & \textbf{0.165 $\pm$ 0.06} & 0.058 $\pm$ 0.01 & \underline{0.151 $\pm$ 0.08} \\
AVATAR          & 0.132 $\pm$ 0.01 & 0.174 $\pm$ 0.03 & 0.064 $\pm$ 0.01 & 0.168 $\pm$ 0.01 \\
TimeGAN         & 0.194 $\pm$ 0.03 & 0.238 $\pm$ 0.09 & \underline{0.051 $\pm$ 0.02} & 0.244 $\pm$ 0.07 \\
T-Forcing       & 0.282 $\pm$ 0.08 & 0.377 $\pm$ 0.11 & 0.134 $\pm$ 0.05 & 0.392 $\pm$ 0.06 \\
Vanilla DDPM    & 0.366 $\pm$ 0.12 & 0.402 $\pm$ 0.14 & 0.214 $\pm$ 0.07 & 0.388 $\pm$ 0.08 \\
\midrule
\multicolumn{5}{c}{\textit{Predictive Score}} \\
\midrule
\textbf{TIMED} & \textbf{0.037 $\pm$ 0.00} & \textbf{0.122 $\pm$ 0.02} & \textbf{0.094 $\pm$ 0.00} & \textbf{0.101 $\pm$ 0.01} \\
DiffTime        & \underline{0.043 $\pm$ 0.01} & \underline{0.136 $\pm$ 0.03} & {0.119 $\pm$ 0.01} & 0.144 $\pm$ 0.02 \\
AVATAR          & 0.044 $\pm$ 0.00 & 0.151 $\pm$ 0.01 & 0.142 $\pm$ 0.03 & \underline{0.122 $\pm$ 0.01} \\
TimeGAN         & 0.048 $\pm$ 0.00 & 0.165 $\pm$ 0.05 & \underline{0.096 $\pm$ 0.01} & 0.165 $\pm$ 0.05 \\
T-Forcing       & 0.058 $\pm$ 0.00 & 0.192 $\pm$ 0.06 & 0.190 $\pm$ 0.07 & 0.314 $\pm$ 0.15 \\
Vanilla DDPM    & 0.168 $\pm$ 0.09 & 0.299 $\pm$ 0.09 & 0.382 $\pm$ 0.12 & 0.335 $\pm$ 0.16 \\
\bottomrule
\end{tabular}
\end{table*}

\begin{itemize}
    \item \textbf{Discriminative Score.} A GRU classifier is trained to distinguish real from synthetic sequences, with real data labeled as ``real'' and generated data as ``synthetic.'' The test accuracy reflects distinguishability; we report \( 0.5 - \text{error} \) for interpretability, where 0 indicates perfect similarity.

    \item \textbf{Predictive Score.} A GRU model is trained on synthetic data to forecast future steps and evaluated on real data using mean absolute error (MAE). Lower MAE indicates better preservation of temporal dependencies.

    \item \textbf{Distributional Alignment.} We use t-SNE~\cite{tsne} and PCA~\cite{pca} to project sequences into 2D for visual comparison. Closer overlap between real and synthetic distributions implies better approximation of \( p(X_{1:T}) \).
\end{itemize}

\subsection{Hyperparameters}
We train the \textit{TIMED} framework using the Adam optimizer with a learning rate of $10^{-3}$ and a batch size of 128. The model backbone consists of a 4-layer Transformer with 8 attention heads and a hidden dimensionality of 128. The denoising diffusion process spans 500 steps, governed by a linear noise schedule ranging from $10^{-4}$ to $10^{-1}$. Training proceeds in three phases: the supervisor is pretrained for 100 epochs, the diffusion model for another 200 epochs, followed by 300 epochs of joint training to align all components.

\subsection{Results and Discussion.}
To ensure statistical reliability, we repeat each experiment four times and report the mean and standard deviation of the scores in Tables~\ref{tbl:scores} and~\ref{tbl:ablation}.

\begin{figure}
\centering
\includegraphics[width=0.5\textwidth]{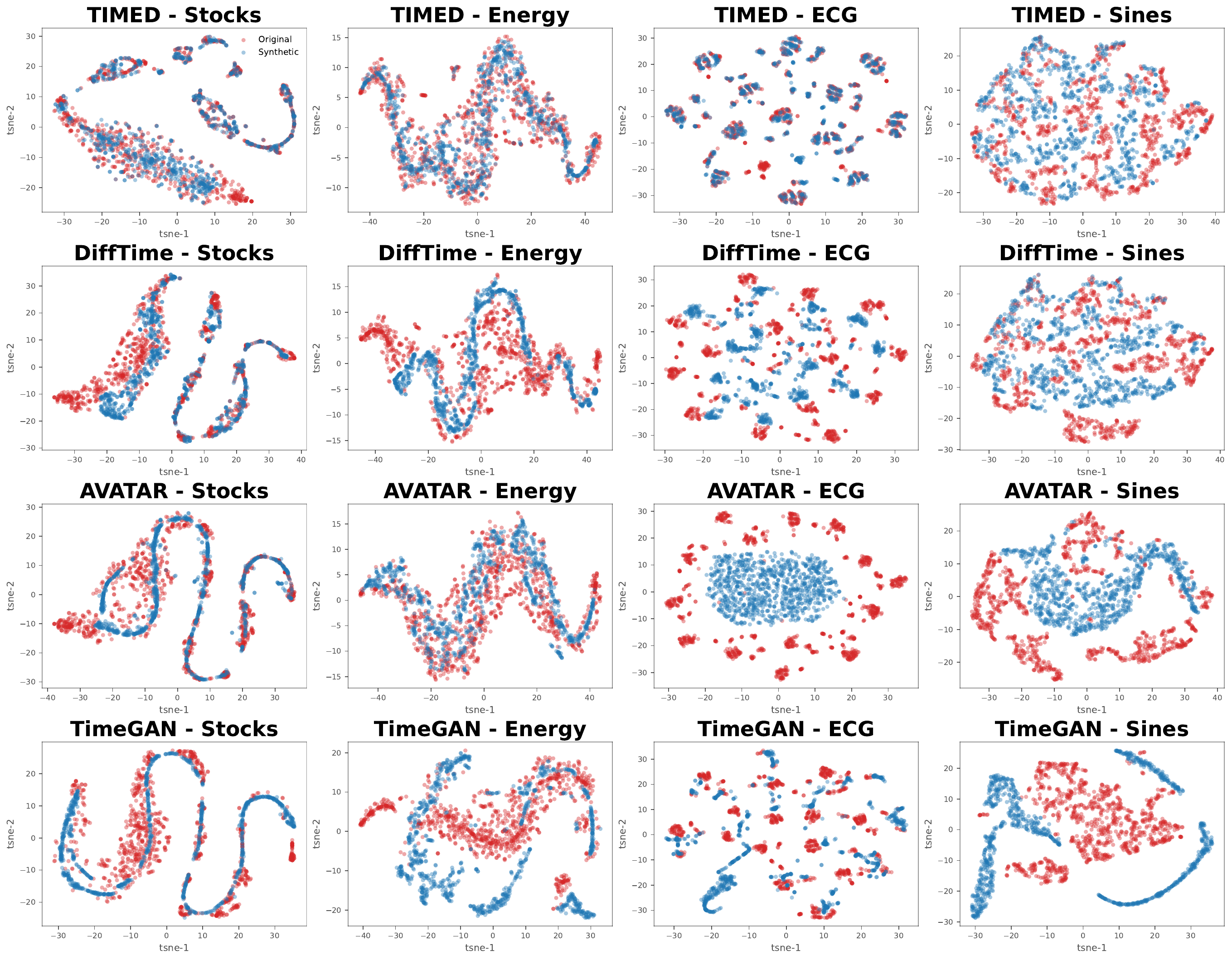}
\caption{t-SNE plots comparing \textit{TIMED} with top-performing baselines across all four datasets. The red markers denote real time series, while blue markers represent generated sequences.}
\label{fig:tsne}
\end{figure}

\begin{figure}
\centering
\includegraphics[width=0.5\textwidth]{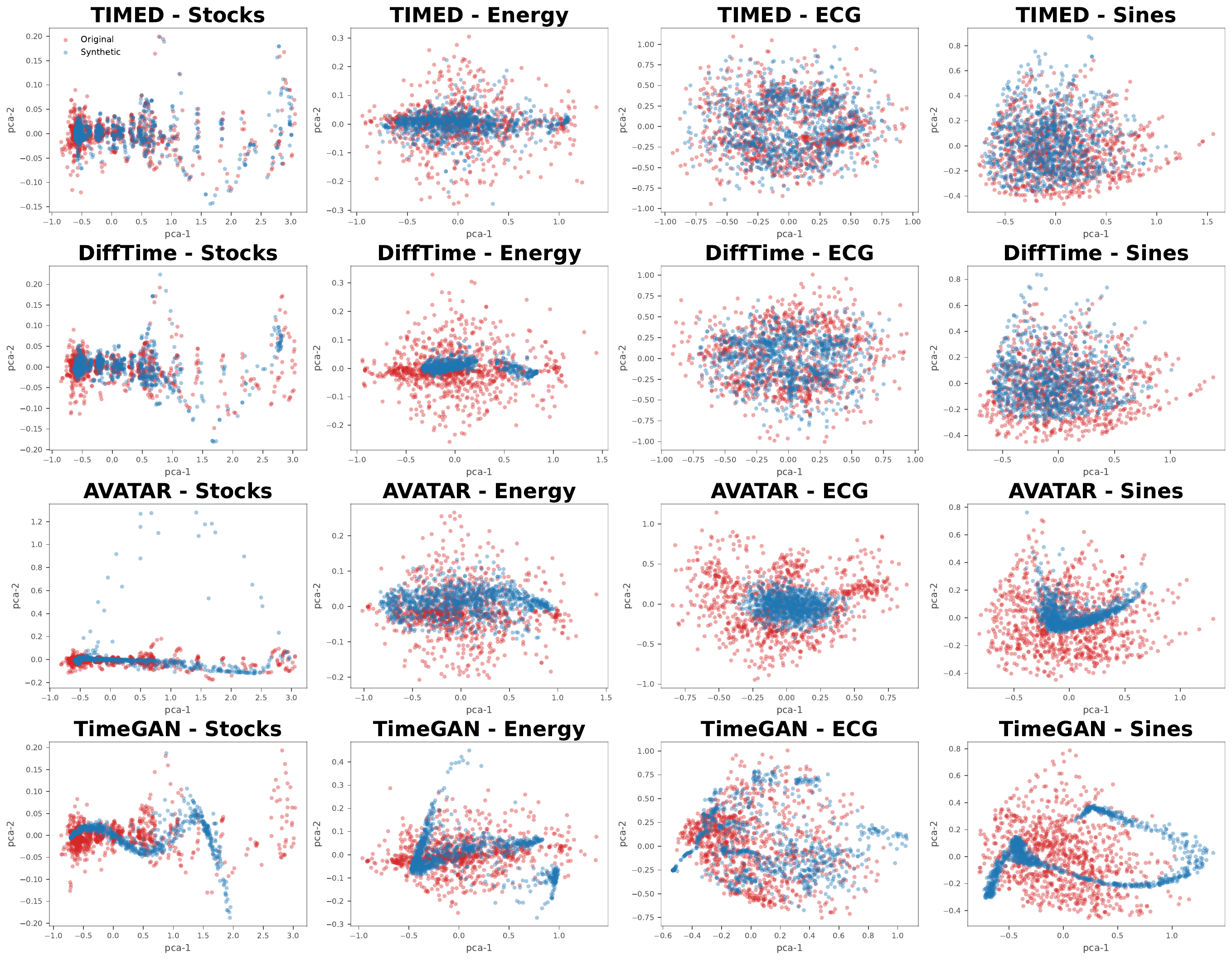}
\caption{PCA visualizations for \textit{TIMED} and the best-performing baselines across the four datasets. Red indicates real samples, whereas blue corresponds to generated data.}
\label{fig:pca}
\end{figure}

As shown in Table~\ref{tbl:scores}, \textit{TIMED} consistently achieves the lowest discriminative and predictive scores across all four benchmark datasets, demonstrating its superior ability to generate realistic and temporally coherent time series. Compared to DiffTime, AVATAR, and TimeGAN, \textit{TIMED} reduces the average discriminative score by approximately 28\%, 34\%, and 46\%, respectively, and lowers the average predictive score by roughly 21\%, 28\%, and 35\%. These improvements highlight \textit{TIMED}’s effectiveness in capturing both marginal distributions and conditional temporal structure. In contrast, the vanilla DDPM baseline performs the worst, with the highest error margins on all datasets. This degradation stems from its CNN-based U-Net backbone, which lacks the inductive bias required for modeling temporal dependencies and is thus poorly suited for sequential data. Overall, \textit{TIMED}’s architectural design and training strategy yield significant gains in both generation quality and downstream predictive performance.

Based on the PCA and t-SNE visualizations in Figs.~\ref{fig:tsne} and~\ref{fig:pca}, \textit{TIMED} achieves the most accurate distributional alignment between real and synthetic samples across all four datasets. In both projection spaces, samples generated by \textit{TIMED} densely cover the same regions as the real data, indicating strong global structure preservation and high-fidelity generation. Compared to baselines, AVATAR performs second best on \textit{Energy}, while DiffTime shows relatively strong alignment on \textit{Stocks}, \textit{ECG}, and \textit{Sines}. However, all models exhibit degraded performance on \textit{Sines}, likely due to its intricate phase-shifted periodicity. Overall, these visual results, along with the discriminative and predictive score evaluations, highlight \textit{TIMED}’s superior capacity for capturing both global and local structure in time series generation.

\begin{figure}
\centering
\includegraphics[width=0.5\textwidth]{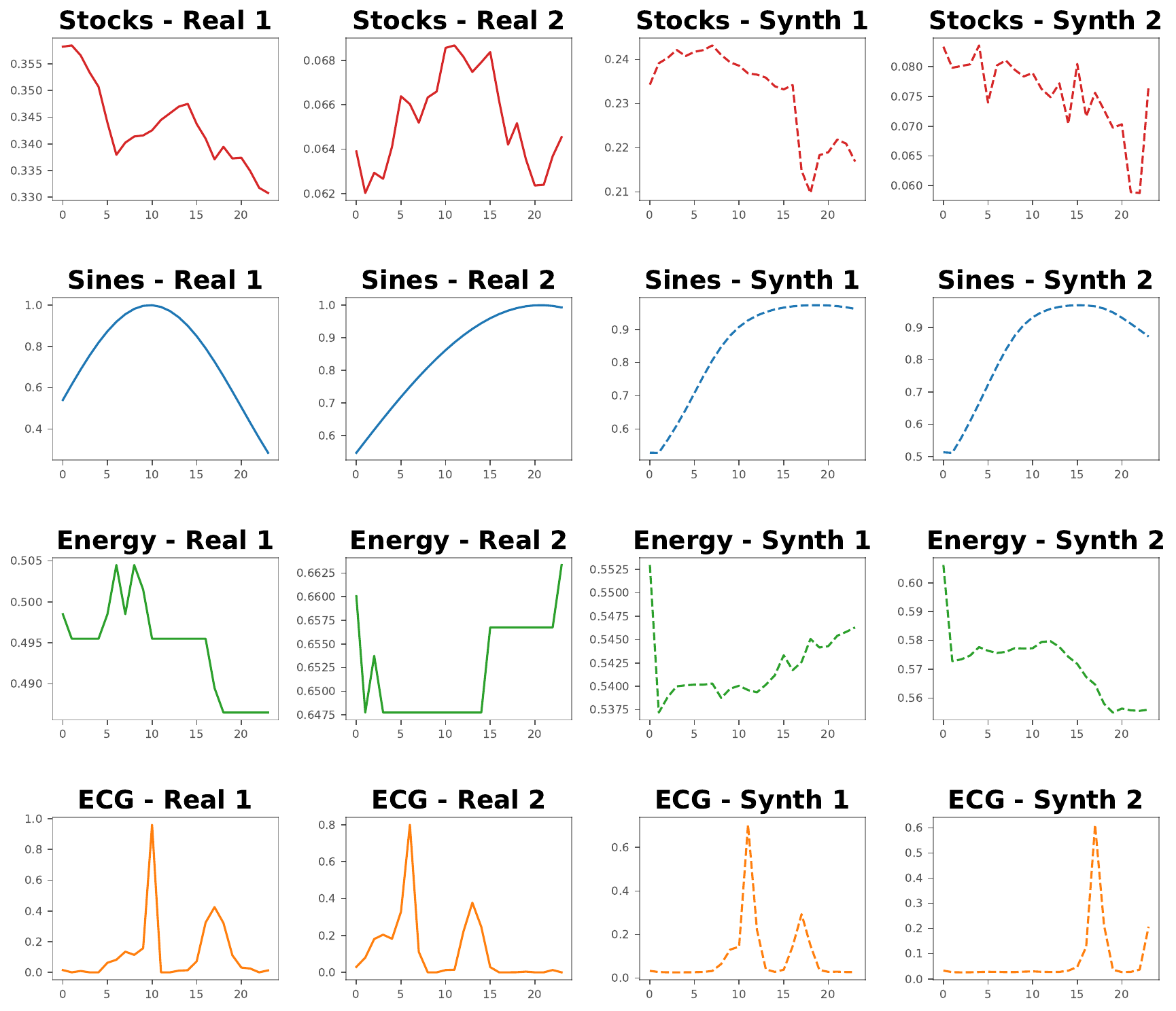}
\caption{Two randomly selected real and synthetic samples (generated by \textsc{TIMED}) are shown for each of the four datasets. The solid lines correspond to original time series, while the dashed lines represent synthetic counterparts.}
\label{fig:example}
\end{figure}

Fig.~\ref{fig:example} presents two real and synthetic sequences per dataset, offering a direct comparison of temporal patterns. The synthetic samples produced by \textit{TIMED} closely mirror the trends, periodicity, and variability observed in the real data, indicating strong sample-level fidelity. This visual alignment supports the model’s ability to capture dataset-specific temporal structure across diverse domains.

\begin{table*}
\centering
\caption{Contribution of each component to the overall performance of the \textit{TIMED} framework. “w/o” denotes the removal of a specific module. “ASL” refers to Autoregressive Supervised Learning, “MMD” indicates the use of Maximum Mean Discrepancy loss, “MA” represents the Masked Attention architecture, and “WC” denotes the Wasserstein Critic network. Lower values indicate better performance.}
\label{tbl:ablation}
\small
\begin{tabular}{lcccc}
\toprule
\multicolumn{5}{c}{\textit{Discriminative Score}} \\ \midrule
\textbf{Variant} & \textbf{Stocks} & \textbf{Sines} & \textbf{ECG} & \textbf{Energy} \\ 
\midrule

\textbf{TIMED} & {0.093 $\pm$ 0.00} & \textbf{0.171 $\pm$ 0.02} & \textbf{0.034 $\pm$ 0.00} & \textbf{0.105 $\pm$ 0.01} \\

w/o ASL & 0.144 $\pm$ 0.01 & 0.290 $\pm$ 0.05 & 0.049 $\pm$ 0.00 & 0.155 $\pm$ 0.02 \\ 

w/o MMD & 0.107 $\pm$ 0.02 & 0.186 $\pm$ 0.04 & 0.056 $\pm$ 0.00 & 0.122 $\pm$ 0.01 \\ 

w/o MA & 0.246 $\pm$ 0.06 & 0.289 $\pm$ 0.09 & 0.106 $\pm$ 0.01 & 0.209 $\pm$ 0.04 \\ 

w/o WC & \textbf{0.091 $\pm$ 0.00} & 0.175 $\pm$ 0.03 & 0.065 $\pm$ 0.00 & 0.110 $\pm$ 0.01 \\

\midrule
\multicolumn{5}{c}{\textit{Predictive Score}} \\ 
\midrule

\textbf{TIMED} & \textbf{0.037 $\pm$ 0.00} & {0.122 $\pm$ 0.02} & \textbf{0.094 $\pm$ 0.00} & \textbf{0.101 $\pm$ 0.01} \\ 

w/o ASL & 0.141 $\pm$ 0.01 & 0.199 $\pm$ 0.03 & 0.179 $\pm$ 0.06 & 0.160 $\pm$ 0.01 \\ 

w/o MMD & 0.044 $\pm$ 0.00 & \textbf{0.120 $\pm$ 0.01} & {0.106 $\pm$ 0.01} & 0.138 $\pm$ 0.01 \\ 

w/o MA & 0.121 $\pm$ 0.01 & 0.182 $\pm$ 0.02 & 0.161 $\pm$ 0.01 & 0.172 $\pm$ 0.04 \\ 

w/o WC & 0.053 $\pm$ 0.00 & 0.128 $\pm$ 0.02 & 0.113 $\pm$ 0.00 & 0.123 $\pm$ 0.05 \\
\bottomrule
\end{tabular}
\end{table*}

\subsection{Ablation Study}
We perform an ablation study to assess the contribution of each novel component in the \textit{TIMED} framework. As shown in Table~\ref{tbl:ablation}, the removal of masked attention (MA) leads to the most significant degradation in performance. Specifically, excluding MA increases the average discriminative score by approximately \textbf{123\%} and the predictive score by \textbf{109\%}, underscoring its importance in modeling temporal dependencies. Similarly, removing the autoregressive supervised loss (ASL) results in a notable performance drop, with a \textbf{63\%} increase in the discriminative score and a \textbf{93\%} increase in the predictive score on average, highlighting its role in enforcing temporal consistency.

In contrast, the Wasserstein Critic (WC) has the least impact among the components. Its removal causes minimal changes in performance, with average increases of less than \textbf{10\%} in both scores. These results demonstrate that while WC offers auxiliary benefits, the primary performance gains in \textit{TIMED} stem from the integration of MA and ASL.

\section{Conclusion}

\textit{TIMED} is a unified generative framework for time series synthesis that integrates a denoising diffusion model with autoregressive supervision and adversarial feedback. It leverages a masked attention backbone across all components to capture both global structure and fine-grained temporal dependencies. A predictive supervisor enhances one-step temporal learning, while a Wasserstein critic promotes temporal coherence and distributional sharpness. Additionally, an MMD loss aligns the generated and real distributions in feature space, improving diversity and fidelity. Empirical results on diverse datasets demonstrate that \textit{TIMED} achieves superior performance over leading baselines, including \textit{DiffTime}, \textit{Vanilla DDPM}, \textit{AVATAR}, and \textit{TimeGAN}. Future work will explore Denoising Diffusion Implicit Models (DDIMs) to accelerate inference and adaptive attention mechanisms to better capture long-range temporal dependencies.

\section*{Acknowledgment}

Shah Muhammad Hamdi acknowledges support from the National Science Foundation (NSF) Geosciences (GEO) Directorate, Division of Atmospheric and Geospace Sciences (AGS), under award \#2301397; the NSF Office of Advanced Cyberinfrastructure (OAC), CRII program, under award \#2305781; and the NSF RISE program, Integrative and Collaborative Education and Research (ICER), under award \#2530946. Soukaina Filali Boubrahimi acknowledges support from the NSF GEO Directorate, AGS, under awards \#2204363 and \#2240022, and from the NSF RISE program, ICER, under award \#2530946.

\balance

\bibliographystyle{IEEEtran}
\bibliography{References}

\end{document}